\documentclass[11pt,oneside,a4paper]{article}

\usepackage[numbers]{natbib}
\let\cite\citep
\usepackage{fullpage}
\usepackage[affil-it]{authblk}
\usepackage{amsmath}
\usepackage{amsthm}
\usepackage{amsfonts}
\usepackage{amssymb}
\usepackage{mathtools}
\usepackage{multirow}
\usepackage{url}
\usepackage{booktabs}
\usepackage{dsfont}
\usepackage{subfigure}
\usepackage{array}
\usepackage{microtype}
\newcommand{\ie}{i.\,e., }
\newcommand{\oms}{\{\!\!\{}
\newcommand{\cms}{\}\!\!\}}

\usepackage{algorithmic}

\newcolumntype{H}{>{\setbox0=\hbox\bgroup}c<{\egroup}@{}}


\title{\textbf{WL-SeqVAE: Sequence-Based Graph Generation Using Variational Autoencoders and Weisfeiler–Leman Colors}}

\author{Iakovos Evdaimon*\textsuperscript{\rm 1},
 Giannis Nikolentzos*\textsuperscript{\rm 2},
 Achilleas Tsimichodimos \textsuperscript{\rm 1},
 Christos Xypolopoulos\textsuperscript{\rm 1,3}, 
 and Michalis Vazirgiannis\textsuperscript{\rm 1,4}\\

\textnormal{\textsuperscript{\rm 1}\'Ecole Polytechnique, IP Paris, France}\\
\textnormal{\textsuperscript{\rm 2}University of Peloponnese, Greece}\\
\textnormal{\textsuperscript{\rm 3} National Technical University of Athens, Greece} \\
\textnormal{\textsuperscript{\rm 4} Mohamed bin Zayed University of Artificial Intelligence, United Arab Emirates}
}
\date{}

\begin{document}
\maketitle 
\begingroup
\renewcommand{\thefootnote}{*}
\footnotetext{These authors contributed equally.}
\endgroup

\begin{abstract}
\noindent In this paper, we introduce WL-SeqVAE, a sequence-based variational autoencoder for generating graphs.
The key observation behind WL-SeqVAE is that node colors induced by the Weisfeiler–Leman (WL) algorithm provide compact discrete summaries of local graph structure and can therefore serve as structural tokens for generation.
Rather than learning a distribution over adjacency matrices directly, our method maps each graph to a sequence of WL colors and employs a variational autoencoder that encodes this sequence into a Gaussian latent distribution, from which an autoregressive decoder reconstructs the graph.
During generation, compatibility constraints derived from the WL colors guide node and edge predictions, combining global information from the latent representation with local structural consistency.
This formulation turns graph generation into a sequence generation problem and avoids directly parameterizing a quadratic adjacency matrix.
We evaluate WL-SeqVAE in the tasks of molecular and general graph generation.
Across the evaluated datasets, WL-SeqVAE is competitive with established baselines on several local structural and molecular quality metrics.
Code is available at \url{https://github.com/iakovosevdaimon/WL-SeqVAE}.
\end{abstract}

\section{Introduction}
In recent years, the availability of graph-structured data has increased significantly across a wide range of application domains, including network science, chemo-informatics, and hardware design.
Unlike conventional data modalities such as vectors or sequences, graphs provide a more expressive representation by explicitly capturing complex relationships and interactions among entities.
For example, in chemo-informatics, molecular structures can be naturally represented as graphs, where nodes correspond to atoms and edges represent chemical bonds.
Graphs thus enable the modeling of interactions that cannot be adequately captured by vector-based representations.

In several domains, there is an increasing demand for generative models capable of producing realistic graphs which preserve the underlying structural characteristics of observed samples~\cite{zhu2022survey}.
For example, in chemo-informatics, graph generative models can produce novel molecular structures with desirable properties, while in network science, synthetic graphs can be used for simulation and privacy-preserving data sharing.
These applications require models that can capture both local connectivity patterns and global structural characteristics of graphs while generating valid and diverse graph instances.

Over the past few years, neural graph generation has become the most widely-used approach for producing synthetic graphs.
Existing methods include autoregressive models~\cite{you2018graphrnn}, variational autoencoders~\cite{simonovsky2018graphvae}, generative adversarial networks~\cite{bojchevski2018netgan}, normalizing flows~\cite{zang2020moflow} and, more recently, diffusion-based methods~\cite{vignac2023digress}.
While existing methods have achieved promising results in different tasks, graph generation remains challenging due to the combinatorial nature of graph structures, permutation invariance, varying graph sizes, and the need to model complex dependencies and interactions among graph nodes~\cite{guo2022systematic}.
Furthermore, the trade-off between generation quality and scalability is a major challenge, which remains largely unresolved.

In this work, we propose WL-SeqVAE, a novel generative model for graph-structured data.
The central novelty of the proposed approach lies in leveraging \textit{node colors} induced by the Weisfeiler–Leman (WL) algorithm~\cite{weisfeiler1968reduction}, a powerful heuristic for graph isomorphism testing.
Unlike conventional graph generative models, which aim to learn a distribution over graph structures, the proposed model learns a distribution over WL-derived node colors directly from observed data.
By shifting the modeling focus from graph topology to color distributions, the proposed framework offers a fundamentally different approach to synthetic graph generation.
We adopt a variational autoencoder (VAE) as the underlying generative model.
An encoder maps the input node sequence to a Gaussian latent distribution, while an autoregressive decoder reconstructs the graph from a sampled latent representation.
Unlike conventional autoregressive graph generators that generate one node at a time, our decoder is guided by \textit{compatibility constraints} derived from the WL colors.
This design combines \textit{global} information encoded in the latent representation with \textit{local} structural consistency enforced by the generated WL colors.
The WL-SeqVAE model is evaluated in the tasks of molecular and general graph generation.
The experimental results demonstrate that WL-SeqVAE is competitive with the current state-of-the-art.
The proposed approach also opens up new opportunities for \textit{scalable graph generation} by casting graph generation as a sequence generation problem, thus mitigating the challenges arising from the quadratic complexity of graph-structured data.

\section{Related Work}\label{sec:related_work}
Graph generation has attracted considerable attention due to applications such as molecular generation~\cite{jin2018junction,shi2020graphaf,zang2020moflow}, protein design~\cite{ingraham2019generative,guo2021generating}, and program synthesis~\cite{brockschmidt2019generative}. Existing approaches can be broadly categorized into six families: autoregressive models, variational autoencoders (VAEs), generative adversarial networks (GANs), normalizing flows, diffusion models, and flow matching.

Autoregressive methods generate graphs sequentially by modeling conditional distributions over nodes and edges.
Representative approaches include GraphRNN~\cite{you2018graphrnn}, GRAN~\cite{liao2019efficient}, GRAM~\cite{kawai2019scalable}, and BiGG~\cite{dai2020scalable}, which improve scalability through sparse representations and efficient decoding.
VAE-based methods learn continuous latent representations from which graphs are reconstructed. Notable examples include GraphVAE~\cite{simonovsky2018graphvae}, CGVAE~\cite{liu2018constrained}, JT-VAE~\cite{jin2018junction}, and extensions for specialized graph families such as DAGs~\cite{zhang2019d,li2020dirichlet,hou2024dag} or other structures~\cite{nikolentzos2023synthetic}.
GAN-based approaches formulate graph generation as an adversarial learning problem.
Representative methods include NetGAN~\cite{bojchevski2018netgan}, which generates random walks, and MolGAN~\cite{de2018molgan}, which combines adversarial learning with reinforcement learning for molecular graph generation.
Normalizing flow methods learn invertible mappings between graph representations and simple base distributions, enabling exact likelihood estimation.
Representative models include GraphNVP~\cite{madhawa2019graphnvp}, GRF~\cite{honda2019graph}, GraphDF~\cite{luo2021graphdf}, and MoFlow~\cite{zang2020moflow}.
Diffusion models have recently become the dominant paradigm for graph generation.
Early methods employed score-based diffusion~\cite{niu2020permutation,jo2022score}, while subsequent work extended the framework to discrete graphs~\cite{vignac2023digress}, autoregressive generation~\cite{kong2023autoregressive}, and latent diffusion in Euclidean and hyperbolic spaces~\cite{yang2024graphusion,evdaimon2024neural,zhou2024unifying,fu2024hyperbolic,wen2024hyperbolic}.
More recently, flow matching has emerged as a computationally efficient alternative to diffusion models.
Existing graph generation methods primarily focus on discrete flow matching, including CatFlow~\cite{eijkelboom2024variational} and Defog~\cite{qin2025defog}, which improve training and sampling efficiency.

The approaches most closely related to ours are motif-based graph generation methods, regardless of the underlying generative framework, which construct graphs from higher-level substructures rather than individual nodes or edges.
Jin et al. proposed in~\cite{jin2020hierarchical} such a hierarchical variational autoencoder that generates molecular graphs from structural motifs.
However, to the best of our knowledge, no prior work has explored the use of the WL algorithm for graph generation.

\section{Preliminaries}\label{sec:preliminaries}
In this section, we first introduce the notation used throughout the paper and then describe the WL algorithm.

\subsection{Notation}
Let $\mathbb{N}$ denote the set of natural numbers. 
Then, $[n] = \{1,\ldots,n\} \subset \mathbb{N}$ for $n \geq 1$.
Let also $\oms \cms$ denote a multiset, \ie a generalized concept of a set that allows multiple instances for its elements.
Let $G = (V,E)$ be an undirected, unweighted graph, where $V$ is the vertex set and $E$ is the edge set.
We will denote by $n$ the number of vertices and by $m$ the number of edges ($n = |V|$ and $m = |E|$).
We will also denote by $\text{id}(v)$ the original node index of node $v$.
Let $\mathcal{N}(v)$ denote the neighborhood of vertex $v$, \ie the set $\{u \mid \{v,u\} \in E\}$.
The degree of a vertex $v$ is $\deg(v) = |\mathcal{N}(v)|$.
The eccentricity $\epsilon(v)$ of a node $v$ is the maximum distance from that node to any other node in the graph.
The $k$-ring of a node consists of all nodes whose shortest path distance from the node is exactly $k$.
Given two nodes $v,u$, $d_{\text{SP}}(v,u)$ denotes the shortest path distance between the two nodes.
An unresolved half-edge (also called a stub) is an edge endpoint that has not yet been matched with another endpoint.
A colored graph is a tuple $(G, c)$, where $G$ is a graph and $c \colon V \rightarrow \Sigma$ is a function that assigns colors (\ie elements from a particular set $\Sigma$) to the vertices of the graph.
We interpret all graphs treated in this paper as colored graphs and just write $G$ instead of $(G,c)$ when $c$ is clear from the context.
If the coloring is not specified, we assume a monochromatic coloring (\ie all vertices have the same color).

\subsection{Weisfeiler-Leman Algorithm}
In recent years, the Weisfeiler–Leman algorithm and its variants have gained considerable interest due to the central role they play in the development of various graph kernels and neural networks~\citep{morris2023weisfeiler}.
We next present the $1$-dimensional Weisfeiler–Leman (WL) algorithm for colored graphs, often referred to as color refinement or naive vertex classification.
The algorithm operates by repeatedly refining a partition of the vertex set according to a prescribed refinement rule until a stable partition is obtained.

Let $\mathcal{G} = \{ G_1, \dots, G_N \}$ denote a collection of graphs where $G_i = (V_i, E_i)$ for $i \in [N]$.
Let also $V$ denote the union of the vertex sets of all graphs, \ie $V = V_1 \cup \dots \cup V_N$.
The WL algorithm runs for a number of iterations, and in each iteration, it computes a vertex coloring $c^{(\ell)} \colon V \to \Sigma^{(\ell)}$ where $\Sigma^{(\ell)}$ denotes the set of colors assigned to vertices during the $\ell$-th iteration of the algorithm.
Specifically, for $\ell \in \mathbb{N}$, the coloring $c^{(\ell)}$ computed by WL after $\ell$ iterations on $\mathcal{G}$ is defined as $c^{(\ell)}(v) = \big(c^{(\ell-1)}(v), \oms c^{(\ell-1)}(w) \mid w \in N(v) \cms \big)$.
That is, $c^{(\ell)}(v)$ consists of the color of $v$ from the previous iteration as well as the multiset of colors of neighbors of $v$ from the previous iteration.
It is clear that each vertex coloring $c^{(\ell)}$ naturally partitions $V$ into color classes, \ie sets of vertices with the same color.
Since the refinement takes the color $c^{(\ell-1)}(v)$ of a vertex $v$ into account when computing $c^{(\ell)}(v)$, the implication $c^{(\ell-1)}(v) \neq c^{(\ell-1)}(u) \Rightarrow c^{(\ell)}(v) \neq c^{(\ell)}(u)$ holds for all $u, v \in V$.
Hence, the color classes induced by $c^{(\ell)}$ are at least as fine as those induced by $c^{(\ell-1)}$.
Let $c \in \Sigma^{(\ell)}$.
We also denote by $\text{Pa}(c)$ the unique parent color of $c$, \ie the color in $\Sigma^{(\ell-1)}$ from which $c$ is derived.
Similarly, we denote by $\text{Ch}(c)$ the set of child colors of $c$, \ie $\text{Ch}(c) = \{ c' \in \Sigma^{(\ell+1)} \colon c' \text{ is derived from } c \}$.

\section{The WL-SeqVAE Graph Generator}\label{sec:methodology}
\begin{figure*}[t]
    \centering
    \includegraphics[width=\textwidth]{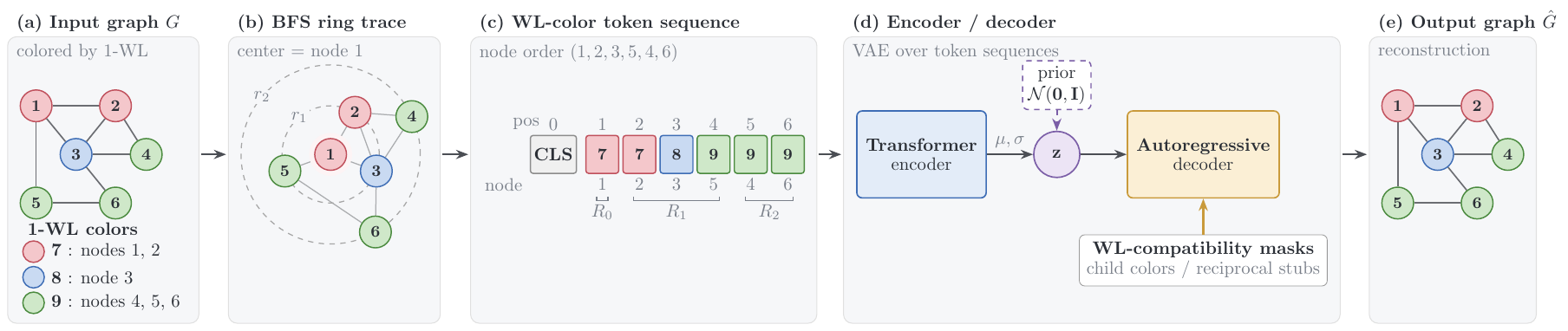}
    \caption{\textbf{Overview of WL-SeqVAE.}
    A BFS ring-ordered sequence of 1-WL color tokens is encoded into a Gaussian latent representation \(\mathbf{z}\). Conditioned on \(\mathbf{z}\), a ring-wise autoregressive decoder generates the graph using WL-compatibility constraints.}
    \label{fig:wl-seqvae}
\end{figure*}

Our model is built on a VAE framework.
The encoder maps the input \textit{sequence of node colors} to a latent distribution, modeled as a multivariate Gaussian parameterized by a mean vector and a standard deviation vector.
A latent representation is then obtained by sampling from this distribution using the reparameterization trick.
Finally, the decoder reconstructs the input sequence from this latent embedding.
An overview of the proposed model is provided in Figure~\ref{fig:wl-seqvae}.
We next present the main components of the proposed model.

\subsection{Node colors}
The node colors are obtained by applying the WL algorithm to the graphs of the training set for $L$ iterations.
Starting from an initial coloring, the WL procedure iteratively updates the color of each node by aggregating the colors of its neighboring nodes and assigning a new color based on the resulting neighborhood signature.
After $L$ iterations, the final node colors encode structural information from the $L$-hop neighborhoods of the nodes.
These colors are then used as node representations.
The model uses a separate embedding layer for the colors produced at each WL iteration.
We denote by $\text{E}_{\Sigma^{(\ell)}}$ the embedding layer for the colors produced at the $\ell$-th WL iteration.

\subsection{Encoder}
The encoder is implemented as a stack of Transformer encoder layers.
Each layer follows the standard Transformer structure with multi-head self-attention and feedforward layers.
A graph $G=(V,E)$ is represented as a sequence of node colors generated by the WL algorithm.
It should be noted that the nodes are not ordered arbitrarily when constructing this sequence.
Instead, the ordering begins by selecting a \textit{central node}, defined as a node with minimum eccentricity.
If multiple nodes satisfy this criterion, ties are broken by selecting the node with the highest degree and, if necessary, the smallest node index.
Let $v$ denote the selected central node.
For each remaining node $u \in V \setminus \{ v \}$, we construct the \textit{signature} $\big(d_{\mathrm{SP}}(v,u), -\deg(u), \mathrm{id}(u)\big)$.
The remaining nodes are then ordered lexicographically according to these signatures.
A special [CLS] token is also prepended to the input sequence.
Specifically, given an input sequence of node colors for $n$ nodes ($n+1$ tokens because of the [CLS] token), token embeddings are first computed as:
\begin{equation*}
  \mathbf{h}_t^{(0)} = \text{E}_{\Sigma^{(L)}} \Big(c_t^{(L)} \Big)+\text{PE}(t)
\end{equation*}
where $\text{PE}(\cdot)$ denotes the positional encoding of the input.
Each encoder layer applies pre-normalization followed by self-attention:
\begin{equation*}
  \tilde{\mathbf{h}}_t^{(\ell)} = \mathbf{h}_t^{(\ell-1)} + \text{MHA} \big(\text{LN} \big(\mathbf{h}_t^{(\ell-1)} \big) \big)
\end{equation*}
and a feedforward network:
\begin{equation*}
  \mathbf{h}_t^{(\ell)} = \tilde{\mathbf{h}}_t^{(\ell)} + \text{FFN} \big(\text{LN} \big(\tilde{\mathbf{h}}_t^{(\ell)} \big) \big)
\end{equation*}
where $\text{LN}(\cdot)$ denotes layer normalization~\cite{ba2016layer}.
Let $\mathbf{H}^{(\ell)}$ denote a matrix whose rows store the normalized representations of the different tokens, \ie its $i$-th row stores $\mathbf{h}_i^{(\ell)}$.
The multi-head attention mechanism is then  defined as:
\begin{equation*}
  \text{MHA}(\mathbf{H}^{(\ell)}) = \big[ \text{head}_1(\mathbf{H}^{(\ell)}) \| \dots \| \text{head}_H(\mathbf{H}^{(\ell)}) \big] \mathbf{W}^{O}
\end{equation*}
where $\big[ \cdot \| \dots \| \cdot \big]$ denotes concatenation and attention heads are defined as:
\begin{equation*}
  \text{head}_i(\mathbf{H}^{(\ell)}) = \text{softmax} \bigg( \frac{\mathbf{H}^{(\ell)} \mathbf{W}_i^{(Q)} \big(\mathbf{H}^{(\ell)} \mathbf{W}_i^{(K)} \big)^\top}{\sqrt{d}_k} \bigg) \mathbf{H}^{(\ell)} \mathbf{W}_i^{(V)}
\end{equation*}
The feedforward network uses GELU activations and expands the hidden dimension to $4d_h$:
\begin{equation*}
  \text{FFN}(\mathbf{h}) = \mathbf{W}_2 \, \text{GELU}(\mathbf{W}_1 \mathbf{h} + \mathbf{b}_1) + \mathbf{b}_2
\end{equation*}
where $\mathbf{W}_1 \in \mathbb{R}^{4d_h \times d_h}$ and $\mathbf{W}_2 \in \mathbb{R}^{d_h \times 4d_h}$.
Dropout is applied throughout the layers for regularization.

Once the final token representations are produced, the vector of the [CLS] token is normalized and projected to Gaussian parameters:
\begin{equation*}
    \begin{split}
        \mathbf{h}_{\text{[CLS]}} &= \text{LN} \big(\mathbf{h}_1^{(L)} \big) \\
        \boldsymbol{\mu}_{\phi} &= \mathbf{W}_{\mu} \mathbf{h}_{\text{[CLS]}} + \mathbf{b}_{\mu} \\
        \log \boldsymbol{\sigma}_{\phi}^2 &= \mathbf{W}_{\log} \mathbf{h}_{\text{[CLS]}} + \mathbf{b}_{\log} \\
    \end{split}
\end{equation*}
During training, the reparameterization trick is utilized:
\begin{equation*}
  \mathbf{z} = \boldsymbol{\mu}_{\phi} + \boldsymbol{\epsilon} \odot \exp \bigg( \frac{1}{2} \log \boldsymbol{\sigma}_{\phi}^2 \bigg), \qquad \boldsymbol{\epsilon} \sim \mathcal{N}(\mathbf{0},\mathbf{I})
\end{equation*}
where $\mathbf{0}$ and $\mathbf{I}$ denote the zero vector and identity matrix, respectively.
During evaluation of the reconstruction loss, we set $\mathbf{z}$ to the mean of the distribution, \ie $\mathbf{z} = \boldsymbol{\mu}_{\phi}$.
For generation from the model prior, the model samples $\mathbf{z} \sim \mathcal{N}(\mathbf{0},\mathbf{I})$.

\subsection{Decoder}
The decoder first transforms the sampled latent vector $\mathbf{z}$ that represents a sequence of node colors as follows:
\begin{equation*}
    \mathbf{h}_z = f_z(\mathbf{z})
\end{equation*}
where $f_z$ is a two-layer MLP with SiLU activation function.

Generation proceeds in a \textit{ring-by-ring} manner, starting from the central node $v$ and subsequently generating nodes belonging to the $k$-rings for $k=1,2, \ldots,\epsilon(v)$.
Specifically, the decoder first predicts the final WL color of the central node.
The prediction probabilities are computed as:
\begin{equation*}
    \mathbf{p}_v = f_{\text{center}}(\mathbf{h}_z + \mathbf{h}_b)
\end{equation*}
where $\mathbf{h}_b$ is a vector that encodes information about the generation state and is defined as:
\begin{equation*}
    \mathbf{h}_b = f_b \bigg( \Big[\text{E}_n(n^*) \, \big\| \, \text{E}_n(n_{\text{cur}}) \, \big\| \, \text{E}_n(n_{\text{rem}}) \, \big\| \, \text{E}_n(u_{\text{stub}}) \, \big\| \, \text{E}_r(k) \Big] \bigg) 
\end{equation*}
where $n^*$ is the target number of nodes, $n_{\text{cur}}$ is the number of nodes that have already been created, $n_{\text{rem}}$ is the remaining number of nodes to be generated (\ie $n_{\text{rem}} = n^* - n_{\text{cur}}$), $u_{\text{stub}}$ is the total number of unresolved half-edges that have emerged from the WL colors of the nodes that have already been generated, and $k$ denotes the current ring index (\ie the ring from which nodes are currently being generated).
All the above quantities are scalars and are embedded to vectors using embedding layers $\text{E}_n$ and $\text{E}_r$, and then concatenated to form a single vector.

Given the final WL color of a node $v$, the penultimate WL colors of all its neighbors are \textit{uniquely determined}.
Consider a neighbor $u$ of $v$.
Let $c^{(L-1)}(u)$ denote the required penultimate WL color of $u$.
The logits corresponding to the node $u$'s final WL color are then given by:
\begin{equation*}
    \mathbf{p}_u = f_{\text{color}} \bigg( \Big[\mathbf{h}_z + \mathbf{h}_b \, \Big\| \, \text{E}_{\Sigma^{(L-1)}}\Big(c^{(L-1)}(u) \Big) \, \Big\| \, \text{E}_r \big( d_{\text{SP}}(v,u) \big) \, \Big\| \, \mathbf{h}_G + \mathbf{h}_b \Big] \bigg)
\end{equation*}
where $\mathbf{h}_G$ is a graph-level representation summarizing the graph constructed so far.
Vector $\mathbf{h}_G$ is defined as:
\begin{equation*}
    \mathbf{h}_G = \frac{1}{n_{\text{cur}}} \sum_{v \in V_{\text{cur}}} \mathbf{h}_v
\end{equation*}
where $V_{\text{cur}}$ denotes the set of nodes that have already been created, and $\mathbf{h}_v$ denotes a vector that encodes information about node $v$, defined as follows:
\begin{equation*}
\begin{aligned}
\mathbf{h}_u
= f_{\text{node}}\bigg( \Big[
    \text{E}_{\Sigma^{(L)}}\Big( c^{(L)}(u) \Big)
    \, \Big\| \, \text{E}_{\Sigma^{(L-1)}} \Big(c^{(L-1)}(u) \Big) \, \Big\| \, \text{E}_{\Sigma^{(0)}} \Big(c^{(0)}(u) \Big)
    \, \Big\| \, \text{E}_r \big( d_{\text{SP}}(v,u) \big)
    \, \Big\| \, \text{E}_d \big( \text{deg}(u) \big) \Big] \bigg)
\end{aligned}
\end{equation*}
where $E_d$ is an embedding layer for node degrees.
Once the final color $c^{(L)}(u)$ of node $u$ has been determined, the node may be connected to nodes in either the \textit{previous ring} or the \textit{current ring} if compatible unresolved stubs remain available.
For a candidate edge of type $e$ between existing nodes $u$ and $w$, the edge probability is computed as:
\begin{equation*}
    p_e = \sigma \bigg( f_{\text{edge}} \Big( \big[ \mathbf{h}_z + \mathbf{h}_b \, \| \, \mathbf{h}_u \, \| \, \mathbf{h}_w \, \| \, \text{E}_e(e) \big] \Big) \bigg) 
\end{equation*}
where $\text{E}_e$ denotes an embedding layer for edge types, and $\sigma$ is the sigmoid activation function.
Note again that a pair of nodes is considered a valid candidate for connection only if it possesses \textit{reciprocal unresolved stubs} that are compatible with edge type $e$.

\subsection{Training Objective}
The loss function of the proposed variational autoencoder has two main terms:
\begin{equation*}
    \mathcal{L} = \mathcal{L}_{\text{rec}} + \beta \mathcal{L}_{\text{KL}} 
\end{equation*}
The first term is the reconstruction loss which measures how well the model reconstructs the input data.
The second term of the loss function is a regularization term.

The reconstruction loss consists of three components:
\begin{equation*}
    \mathcal{L}_{\text{rec}} = \mathcal{L}_{\text{nodes}} + \mathcal{L}_{\text{prev}} + \mathcal{L}_{\text{intra}}
\end{equation*}
Note that we employ \textit{teacher forcing}~\cite{williams1989learning}, \ie at each generation step the model is conditioned on the ground-truth partial graph instead of its own previous predictions.
The first component of the reconstruction term $\mathcal{L}_{\text{nodes}}$ corresponds to the prediction of node colors.
For the central node, the model predicts a distribution over all colors in the final WL iteration, and a cross-entropy loss is computed with respect to the ground-truth color.
For all remaining nodes, the prediction is restricted to the children of the node's color in the penultimate WL iteration.
The cross-entropy loss is then computed over this reduced set of candidate colors.
Formally, we have:
\begin{equation*}
    \begin{split}
        \mathcal{L}_{\text{nodes}} = -\sum_{c \in \Sigma^{(L)}} \mathds{1}_{\{c=c^{(L)}(v)\}} \log [\mathbf{p}_v]_c - \frac{1}{n-1} \sum_{u \in V \setminus \{ v \}} \sum_{c \in \text{Ch}(c^{(L-1)} (u))} \mathds{1}_{\{c=c^{(L)}(u)\}} \log [\mathbf{p}_u]_c
    \end{split}
\end{equation*}
where $[\mathbf{p}_v]_c$ denotes the predicted probability that node $v$ is assigned color $c$.
Note that the loss term places greater emphasis on the color of the central node than on the colors of the remaining nodes.
This encourages more accurate color prediction for the central node, which plays the most important role in the generation procedure.

The other two components of the reconstruction term $\mathcal{L}_{\text{rec}}$ are associated with edge prediction.
Specifically, they model ($1$) edges between a node in ring $r$ and nodes in the preceding ring $r-1$; and ($2$) edges between nodes within the same ring.
Both components are defined using binary cross-entropy losses:
\begin{equation*}
    \begin{split}
        \mathcal{L}_{\text{prev}} &= -\frac{1}{|\mathcal{P}|} \sum_{e \in \mathcal{P}} \Big( y_e \log p_e + (1 - y_e) \log(1 - p_e) \Big) \\
        \mathcal{L}_{\text{intra}} &= -\frac{1}{|\mathcal{I}|} \sum_{e \in \mathcal{I}} \Big( y_e \log p_e + (1 - y_e) \log(1 - p_e) \Big) \\
    \end{split}
\end{equation*}
Here, $\mathcal{P}$ denotes the set of candidate edges connecting nodes in ring $r$ to nodes in ring $r-1$, and $\mathcal{I}$ denotes the set of candidate edges between nodes within the same ring.
The binary label $y_e$ indicates whether candidate edge $e$ is present in the input graph or absent, while $p_e$ denotes the predicted probability that edge $e$ exists.

The second term of the loss function measures how closely the output distribution matches $p(\mathbf{z})$, where $p(\mathbf{z}) = \mathcal{N}(\mathbf{0},\mathbf{I})$ is an isotropic prior:
\begin{equation*}
    \mathcal{L}_{\text{KL}} = \text{KL} \big(q_{\phi}(\mathbf{z} \mid \tau) \, \| \, \mathcal{N}(\mathbf{0},\mathbf{I}) \big) = -\frac{1}{2} \sum_{j=1}^{d_z} (1 + \log \boldsymbol{\sigma}_j^2 - \boldsymbol{\mu}_j^2 - \boldsymbol{\sigma}_j^2)
\end{equation*}
In our implementation, we average the above quantity over the batch.

\subsection{Sampling}
At inference time, new graphs are generated according to the following procedure:
\begin{enumerate}
    \item We sample $\mathbf{z} \sim \mathcal{N}(\mathbf{0},\mathbf{I})$ and then compute $\mathbf{h}_z = f_z (\mathbf{z})$.
    \item We sample a target number of nodes $n^*$ from the empirical training distribution and set $n_{\max} = \lceil \gamma n^* \rceil$ where $\gamma$ is a hyperparameter.
    \item We compute the vector $\mathbf{h}_b$ and then produce the final WL color of the central node.
    \item We create the central node, place it in ring $0$, and initialize its unresolved half-edges from its final WL color.
    \item For ring $r = 1,\dots, r_{\max}$:
    \begin{itemize}
        \item[(a)] Collect the unresolved stubs originating from the previous ring.
        \item[(b)] For each source stub, identify compatible existing nodes and sample whether to connect the stub to one of them.
        \item[(c)] If no existing node is selected and $n_{\text{cur}} < n_{\max}$, sample a compatible child final WL color, create the corresponding node in the current ring, and connect it to the source node.
        \item[(d)] After processing the source stubs, predict additional edges between current-ring nodes with compatible unresolved stubs.
        \item[(e)] Once all stubs originating from the previous ring have been processed, advance to the next ring.
        \item[(f)] Terminate when all stubs have been resolved or the maximum number of nodes, $n_{\max}$, has been reached.
    \end{itemize}
\end{enumerate}
If generation terminates with no remaining stubs for every generated node, then each node realizes the full neighbor multiset required by its final WL color.
We call such a graph \textit{WL-complete}.
During generation, the procedure may terminate after reaching the maximum number of nodes $n_{\max}$ while some nodes still contain unresolved stubs.

\section{Experimental Evaluation}\label{sec:experimental_evaluation}
We evaluate WL-SeqVAE in two main tasks.
First, we consider general graph generation to determine whether the model captures the structural properties of different graph families. 
We then evaluate the model on two widely used molecular benchmarks.
These experiments assess whether the model can generate chemically valid and diverse molecular graphs. 

\paragraph{Model Selection.}
For each dataset, we train three different models by setting the number of WL iterations to $1$, $2$, and $3$. 
For each one of the three models, we store two candidate checkpoints based on validation data: 
($1$) the \emph{best validation loss} checkpoint corresponds to the epoch with the lowest validation objective under the epoch-specific KL warm-up weight; and
($2$) the \emph{best structural} checkpoint instead minimizes a validation generation score that penalizes failed or invalid generated graphs, duplicates, training-set matches, incomplete WL constructions, and unresolved half-edges. 
It also includes the mean of the log-scaled degree, clustering, and orbit $\mathrm{MMD}^2$ values. 
We compare all combinations of WL iterations and checkpoint types on the validation set and select a single configuration before evaluating on the test set.

Table~\ref{tab:wl-selection-summary} reports the selected configuration for each dataset. 
It also reports the WL-completeness for each dataset, defined as the fraction of generated samples that have all local WL constraints satisfied and no unresolved stubs. 
This metric evaluates whether the decoder completes its WL-based construction consistently.
The complete record for every evaluated WL depth, including the number of epochs, the batch size and the total training time is provided in the Appendix.
The Appendix also reports the remaining WL depths and checkpoint roles, together with the complete training configuration, sampling protocol, and decoder controls.

\begin{table}[t]
\centering
\footnotesize
\setlength{\tabcolsep}{2.5pt}
\renewcommand{\arraystretch}{1.05}
\begin{tabular}{@{}l|c|c|c}
\toprule
\multirow{2}{*}{Dataset} & WL  & Selected & \multirow{2}{*}{WL-Complete (\%) \(\uparrow\)} \\
& iteration & checkpoint \\
\midrule
Ego-small       & 1 & best val. loss & 89.17 $\pm$ {\scriptsize 4.71 } \\
Community-small & 3 & best structural  & 20.00 $\pm$ {\scriptsize 7.07} \\
Enzymes         & 3 & best val. loss & 31.91 $\pm$ {\scriptsize 0.40} \\
Grid            & 3 & best val. loss & 11.67 $\pm$  {\scriptsize 8.50} \\
QM9             & 2 & best val. loss & 54.85 $\pm$  {\scriptsize 0.83} \\
ZINC250K        & 2 & best structural & 57.47 $\pm$ {\scriptsize 1.43} \\
\bottomrule
\end{tabular}
\caption{Selected WL-SeqVAE configurations based on the validation set.
Completion is the percentage of samples that are WL-complete and contain no unresolved WL stubs.
We report the mean and standard deviation across three different seeds.}
\label{tab:wl-selection-summary}
\end{table}

\subsection{General Graph Generation}
Following the evaluation protocol of CatFlow~\cite{eijkelboom2024variational}, we first evaluate the proposed method on the following $4$ synthetic and real-world datasets: ($1$) Ego-small; ($2$) Community-small; ($3$) Enzymes; and ($4$) Grid.
For all datasets, we use the ordered GDSS splits~\cite{jo2022score}. 
For each experiment (seed and dataset), we generate the same number of graphs as in the corresponding test set, Specifically, we generate $40$ graphs for Ego-small, $20$ for Community-small, $117$ for Enzymes, and $20$ for Grid.

\paragraph{Experimental setup.}
We measure the agreement between the generated and test distributions using the squared maximum mean discrepancy ($\mathrm{MMD}^2$).
Degree and clustering distributions are compared using Gaussian--Earth Mover's Distance kernels, while the orbit metric uses a Gaussian kernel over exact ORCA node-orbit count vectors for graphlets of up to four nodes.
Lower values indicate closer agreement with the test distribution.

\paragraph{Baselines.}
We compare WL-SeqVAE against the following $5$ baselines: ($1$) GraphVAE~\cite{simonovsky2018graphvae}, ($2$) GNF~\cite{liu2019graph}; ($3$) EDP-GNN~\cite{niu2020permutation}; ($4$) GDSS~\cite{jo2022score}; and ($5$) CatFlow~\cite{eijkelboom2024variational}.
We use the results reported in~\cite{eijkelboom2024variational}. 
The full baseline set is available for Ego-small and Community-small. 
For Enzymes and Grid, we compare only against CatFlow.

\paragraph{Results.}
The results are shown in Table~\ref{tab:abstract-graph-baselines}.
On Ego-small, WL-SeqVAE outperforms GraphVAE, GNF, and EDP-GNN in degree and clustering, with a degree score close to that of GDSS.
Its orbit score of $0.008$ matches the CatFlow mean, although GNF, EDP-GNN, and GDSS achieve slightly better results on this metric.
On Community-small, WL-SeqVAE achieves a degree $\mathrm{MMD}^2$ of $0.019$, comparable to CatFlow $(0.018)$ and lower than those of the other baselines.
It also improves upon GraphVAE, GNF, and EDP-GNN in orbit statistics, but trails GDSS and CatFlow in both clustering and orbit. 
On Enzymes, CatFlow outperforms WL-SeqVAE across all three statistics.
On Grid, WL-SeqVAE achieves a lower orbit $\mathrm{MMD}^2$ than CatFlow
$(0.0442$ vs. $0.075$), but performs less favorably on degree and mainly on clustering. 
Overll, these results suggest that the WL-based representation captures local degree and motif statistics effectively, while graph families with stronger global structure remain more challenging.

\begin{table}[t]
\centering
\footnotesize
\setlength{\tabcolsep}{3pt}
\renewcommand{\arraystretch}{1.05}
\begin{tabular}{lcccccccccccc}
\toprule
& \multicolumn{3}{c}{Ego-small}
& \multicolumn{3}{c}{Community-small} \\
\cmidrule(lr){2-4}
\cmidrule(lr){5-7}
Method
& Degree $\downarrow$ & Clustering $\downarrow$ & Orbit $\downarrow$
& Degree $\downarrow$ & Clustering $\downarrow$ & Orbit $\downarrow$ \\
\midrule

GraphVAE
& 0.130 & 0.170 & 0.050
& 0.350 & 0.980 & 0.540 \\

GNF
& 0.030 & 0.100 & \textbf{0.001}
& 0.200 & 0.200 & 0.110 \\

EDP-GNN
& 0.052 & 0.093 & 0.007
& 0.053 & 0.144 & 0.026 \\

GDSS
& 0.021 & \textbf{0.024} & 0.007
& 0.045 & \textbf{0.086} & \textbf{0.007} \\

CatFlow
& \textbf{0.013} $\pm$ 0.007 & \textbf{0.024} $\pm$ 0.009 & 0.008 $\pm$ 0.005
& \textbf{0.018} $\pm$ 0.012 & \textbf{0.086} $\pm$ 0.021 & \textbf{0.007} $\pm$ 0.005 \\

\textbf{WL-SeqVAE}
& 0.023 $\pm$ 0.012 & 0.033 $\pm$ 0.015 & 0.008 $\pm$ 0.003
& 0.019 $\pm$ 0.011 & 0.183 $\pm$ 0.060 & 0.023 $\pm$ 0.010 \\
\bottomrule
\end{tabular}

\vspace{.1cm}

\begin{tabular}{lcccccc}
\toprule
& \multicolumn{3}{c}{Enzymes}
& \multicolumn{3}{c}{Grid} \\
\cmidrule(lr){2-4}
\cmidrule(lr){5-7}
Method
& Degree $\downarrow$ & Clustering $\downarrow$ & Orbit $\downarrow$
& Degree $\downarrow$ & Clustering $\downarrow$ & Orbit $\downarrow$ \\
\midrule
CatFlow
& \textbf{0.013} $\pm$ 0.012 & \textbf{0.062} $\pm$ 0.011 & \textbf{0.008} $\pm$ 0.007
& \textbf{0.115} $\pm$ 0.010 & \textbf{0.004} $\pm$ 0.002 & 0.075 $\pm$ 0.071 \\

\textbf{WL-SeqVAE}
& 0.029 $\pm$ 0.004 & 0.571 $\pm$ 0.047 & 0.039 $\pm$ 0.012
& 0.132 $\pm$ 0.041 & 0.117 $\pm$ 0.078 & \textbf{0.044} $\pm$ 0.016 \\

\bottomrule
\end{tabular}
\caption{\textbf{Performance on general graph generation}. Baseline results are taken from~\cite{eijkelboom2024variational}. WL-SeqVAE results are reported as the mean $\pm$ standard deviation over three random seeds. MMD values are reported as raw biased \(\mathrm{MMD}^2\); lower is better.}
\label{tab:abstract-graph-baselines}
\end{table}

\subsection{Molecular Graph Generation}
We also evaluate WL-SeqVAE on heavy-atom QM9~\cite{wu2018moleculenet} and ZINC250K~\cite{sterling2015zinc}, following the molecular preprocessing protocols of CatFlow and GDSS and using the corresponding test references.
Each molecule is represented as a labeled graph, with nodes denoting atom types and edges denoting bond types.
The molecules are kekulized, and hydrogen atoms are removed before canonical SMILES are produced.
For QM9, the preprocessing pipeline retains $130,413$ of the $133,885$ raw molecular records.
The remaining $3,472$ records are excluded because they contain invalid molecules, unsupported atom types, or too few heavy atoms.
The resulting graphs contain four atom types: $C$, $N$, $O$ and $F$.
For ZINC250K, the initial split comprises $220,011$ training and $24,445$ validation graphs.
Canonical WL trace construction retains all training graphs and $23,546$ validation graphs, while $899$ validation graphs are excluded because they contain out-of-vocabulary WL colors.
Further preprocessing details are provided in the Appendix.

\paragraph{Evaluation protocol.}
For each seed, we perform $10,000$ independent generation attempts.
Validity is computed over all attempts without valency correction or edge resampling, while uniqueness is computed over the RDKit-valid molecules.
Fréchet ChemNet Distance (FCD) measures the distributional similarity between the valid generated molecules and the test reference set in ChemNet representation space.
Generated duplicates are retained when computing FCD, and invalid generations are not replaced by additional samples.
Following the CatFlow evaluation protocol, we omit novelty because QM9 covers a constrained space of small molecules, while molecular generators typically achieve nearly $100\%$ novelty on ZINC250K~\cite{eijkelboom2024variational}.

\paragraph{Baselines.}
\noindent\textbf{Baselines.}
The proposed approach is compared against the following $9$ baselines: ($1$) MoFlow~\cite{zang2020moflow}; ($2$) EDP-GNN~\cite{niu2020permutation}; ($3$) GraphEBM~\cite{liu2021graphebm}; ($4$) GDSS~\cite{jo2022score}; ($5$) DiGress~\cite{vignac2023digress}; ($6$) Flow Matching~\cite{lipman2023flow}; ($7$) Dirichlet FM~\cite{stark2024dirichlet}; ($8$) CatFlow~\cite{eijkelboom2024variational}; and ($9$) DeFoG~\cite{qin2025defog}.

\paragraph{Results.}
On QM9, WL-SeqVAE achieves \(99.74\%\) validity, showing that the decoder generates chemically valid molecules in almost all attempts. 
It also obtains \(94.74\%\) uniqueness and an FCD of \(1.4071\). 
Although its validity is comparable to the strongest baselines, the lower uniqueness and higher FCD indicate that the generated distribution is less diverse and does not match the test distribution as closely as CatFlow.

On the larger and more diverse ZINC250K dataset, WL-SeqVAE achieves \(93.16\%\) validity and \(100.00\%\) uniqueness, indicating that all valid generated molecules are distinct. 
Its FCD of \(18.1730\) remains competitive with several baselines, although it is higher than the values obtained by GDSS, Dirichlet FM, CatFlow and DeFoG. 
These results show that WL-SeqVAE can generate diverse molecular graphs across datasets with different atom vocabularies and graph-size distributions, while accurately matching the molecular distribution remains more challenging.

The strict WL-completion rates reported in Table~\ref{tab:wl-selection-summary} provide a complementary view of the generation process. 
Despite the high molecular validity, not every generated molecule satisfies all sampled local WL constraints. 
This difference shows that chemical validity and strict WL completion measure distinct properties.
In particular, a molecule may be chemically valid even when the decoder does not fully realize its sampled WL-based construction plan.

\begin{table}[t]
\centering
\footnotesize
\setlength{\tabcolsep}{5pt}
\renewcommand{\arraystretch}{1.05}

\begin{tabular}{@{}lcccccc@{}}
\toprule
& \multicolumn{3}{c}{QM9} & \multicolumn{3}{c}{ZINC250K} \\
\cmidrule(lr){2-4}\cmidrule(l){5-7}
Method & Validity \(\uparrow\) & Uniqueness \(\uparrow\) & FCD \(\downarrow\)
& Validity \(\uparrow\) & Uniqueness \(\uparrow\) & FCD \(\downarrow\) \\
\midrule
MoFlow & 91.36 & 98.65 & 4.467 & 63.11 & 99.99 & 20.931 \\
EDP-GNN & 47.52 & 99.25 & 2.680 & 82.97 & 99.79 & 16.737 \\
GraphEBM & 8.22 & 97.90 & 6.143 & 5.29 & 98.79 & 35.471 \\
GDSS & 95.72 & 98.46 & 2.900 & 97.01 & 99.64 & 14.656 \\
DiGress & 99.00 & 96.20 & -- & -- & -- & -- \\
Flow Matching & 94.10 & 98.20 & 5.155 & 94.01 & 96.68 & 18.764 \\
Dirichlet FM & 99.10 & 98.15 & 0.888 & 97.52 & 99.20 & 14.222 \\
CatFlow & \textbf{99.81} & \textbf{99.95} & 0.441 & 99.21 & \textbf{100.00} & 13.211 \\
DeFoG & 99.4 & 96.3 & \textbf{0.12} & \textbf{99.22} & 99.99 & \textbf{1.425} \\
\midrule
\textbf{WL-SeqVAE}
& 99.74 $\pm$ 0.05 & 94.74 $\pm$ 0.42 & 1.407 $\pm$ 0.079
& 93.16 $\pm$ 1.58 & \textbf{100.00} $\pm$ 0.00 & 18.173 $\pm$ 0.12 \\
\bottomrule
\end{tabular}
\caption{\textbf{Performance on molecular graph generation results under the CatFlow/GDSS protocol}. Baseline results are taken from~\cite{eijkelboom2024variational}. WL-SeqVAE results are reported as the mean $\pm$ standard deviation over three random seeds. Validity and uniqueness are reported as percentages.}
\label{tab:molecular-generation}
\end{table}

\begin{figure}[t]
    \centering
    \includegraphics[width=\textwidth]{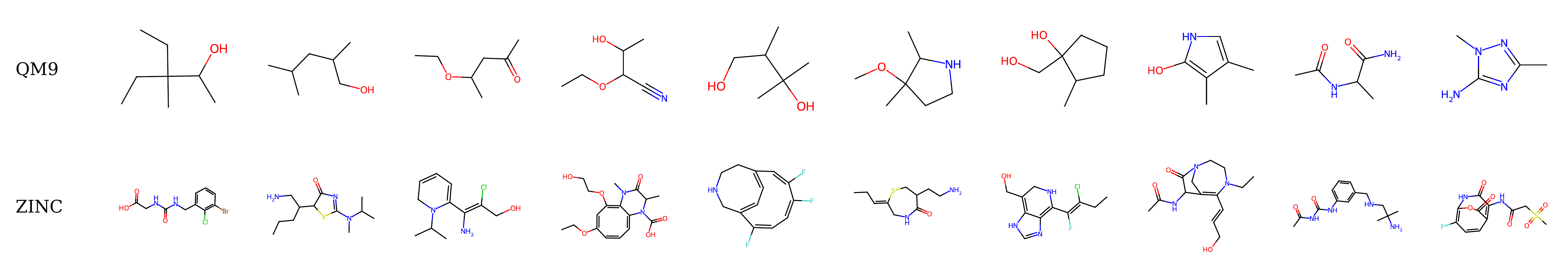}
    \caption{Examples of molecules generated from the WL-SeqVAE model trained on QM9 and ZINC250K.}
    \label{fig:generated_molecules}
\end{figure}

Figure~\ref{fig:generated_molecules} shows representative molecules generated by WL-SeqVAE on QM9 and ZINC.
The samples display structural diversity and are visually consistent with the molecular distributions of the respective datasets.
Samples generated from the model trained on QM9 are small organic molecules with common oxygen- and nitrogen-containing functional groups, while those generated from the mode trained on ZINC are larger drug-like structures featuring aromatic systems, heterocycles, amides, and halogen substituents. Together with the quantitative results reported above, these examples suggest that WL-SeqVAE learns chemically meaningful structural patterns while maintaining diversity across generated molecules.

\section{Conclusion}\label{sec:conclusion}
We presented WL-SeqVAE, a sequence-based VAE that reformulates graph generation as a sequence generation problem by leveraging node colors induced by the WL algorithm as structural tokens.
By encoding sequences of WL colors into a continuous latent representation and decoding them autoregressively under WL-derived compatibility constraints, the proposed framework combines global latent information with local structural consistency while avoiding the direct generation of a quadratic adjacency matrix.
Experimental results on molecular and abstract graph generation demonstrate that WL-SeqVAE is competitive with established baselines across several quality metrics.

\bibliographystyle{plain}
\bibliography{main}

\appendix

\section{Experimental Evaluation}
\label{app:evaluation-protocol}

\subsection{Experimental Setup and Configuration}
\label{app:reproducibility}

\paragraph{Training configuration.}
Unless noted otherwise, we train each model for up to 50 epochs using a hidden dimension of $192$, a latent dimension of $64$, four Transformer layers, four
attention heads, and dropout $0.1$.
We use gradient clipping at $1.0$ and linearly increase the KL weight to $\beta=0.1$ over the first ten epochs.
Table~\ref{tab:sup-optimization-hyperparameters} summarizes the shared optimization settings.
Batch size and maximum sequence length depend on the dataset and WL depth.
Table~\ref{tab:sup-wl-training-summary} reports the total training and validation time for the selected configuration of each dataset, together with the other evaluated WL depths.
The times include all completed training and validation epochs but exclude trace preprocessing, validation generation used by the structural selector, and final test generation.

\begin{table}[t]
\centering
\small
\setlength{\tabcolsep}{10pt}
\renewcommand{\arraystretch}{1.05}
\begin{tabular}{@{}lcc@{}}
\toprule
Hyperparameter & General Graph Generation Datasets & QM9/ZINC250K \\
\midrule
Optimizer     & AdamW & AdamW  \\
Scheduler     & None (constant) & None (constant)  \\
Learning rate & \(3\cdot10^{-4}\) & \(3\cdot10^{-4}\)  \\
Weight decay  & \(10^{-4}\) & \(10^{-4}\) \\

\bottomrule
\end{tabular}
\caption{Optimization hyperparameters used for the general graph generation and molecular generation experiments.}
\label{tab:sup-optimization-hyperparameters}
\end{table}

\begin{table}[t]
\centering
\scriptsize
\setlength{\tabcolsep}{3.5pt}
\renewcommand{\arraystretch}{1.08}
\begin{tabular*}{\textwidth}{@{\extracolsep{\fill}}lccccc@{}}
\toprule
Dataset & \shortstack{WL\\iterations}
& \shortstack{Selected\\checkpoint}
& \shortstack{Total\\epochs}
& \shortstack{Batch\\size}
& \shortstack{Training + validation\\time (min)} \\
\midrule
Ego-small       & 1 & Best val. loss    &  50 & 32 & \(0.34 \pm 0.01\) \\
                & 2 & Best val. loss               & 50 & 32 & \(0.32 \pm 0.00\) \\
                & 3 & Best val. loss              & 50 & 32 & \(0.32 \pm 0.02\) \\
\addlinespace[2pt]
Community-small & 1 & Best structural                      & 50 & 32 & \(0.23 \pm 0.01\) \\
                & 2 & Best structural               & 50 & 32 & \(0.23 \pm 0.01\) \\
                & 3 & Best structural   & 50 & 32 & \(0.23 \pm 0.01\) \\
\addlinespace[2pt]
Enzymes         & 1 & Best val. loss            & 50 & 32 & \(1.42 \pm 0.03\) \\
                & 2 & Best val. loss                & 50 & 32 & \(1.33 \pm 0.01\) \\
                & 3 & Best val. loss    & 50 & 32 & \(1.33 \pm 0.01\) \\
\addlinespace[2pt]
Grid            & 1 & Best val. loss               & 50 & 32 & \(1.03 \pm 0.05\) \\
                & 2 & Best val. loss                   & 50 & 32 & \(0.85 \pm 0.01\) \\
                & 3 & Best val. loss           & 50 & 32 & \(0.72 \pm 0.01\) \\
\addlinespace[2pt]
QM9             & 1 & Best val. loss               & 50 & 256 & \(186.42 \pm 96.74\) \\
                & 2 & Best val. loss    & 50 & 32 & \(277.08 \pm 18.31\) \\
                & 3 & Best val. loss             & 50 & 32 & \(290.45 \pm 3.94\) \\
\addlinespace[2pt]
ZINC250K        & 2 & Best structural   & 50 & 512 & \(356.14 \pm 17.54\) \\
                & 3 & Best structural               & 50 & 256 & \(377.51 \pm 7.71\) \\
\midrule
\multicolumn{6}{@{}l}{\emph{Supplementary results}} \\
\addlinespace[2pt]
PROTEINS        & 1 & Best structural       & 50 & 32 & \(13.28 \pm 0.84\) \\
                & 2 & Best structural   & 50 & 32 & \(13.26 \pm 0.78\) \\
                & 3 & Best structural              & 50 & 16 & \(14.49 \pm 0.10\) \\
\addlinespace[2pt]
Planar          & 3 & Best structural   & 50 & 32 & \(0.51 \pm 0.03\) \\
\addlinespace[2pt]
SBM             & 3 & Best structural   & 50 & 32 & \(0.72 \pm 0.05\) \\
\bottomrule
\end{tabular*}
\caption{Training time across WL depths and the selected configuration for each dataset. Reported times include training and validation and are reported as the mean $\pm$ standard deviation over all seeds.}
\label{tab:sup-wl-training-summary}
\end{table}

\paragraph{Sampling and decoder controls.}
The number of graphs generated for each general graph generation dataset matches the size of its corresponding test split.
For molecular evaluation, each seed yields $10,000$ independent prior-sampling attempts, with invalid generations not replaced.
We use a node-budget factor of $\gamma=1$, giving $n_{\max}=\lceil\gamma n^*\rceil=n^*$.
The sampler prioritizes legal reciprocal connections to existing nodes before creating a compatible child node. The dynamic budget variables defined above guide these decisions but do not enforce an exact graph size or density. Specifically, generation may terminate before reaching $n^*$ nodes, but never exceeds this limit.
We report strict WL completion separately from the number of returned graphs and dataset-specific validity.

\paragraph{Checkpoint roles.}
We select between two checkpoint candidates based on validation data.
As already discussed, the \textit{best validation loss} checkpoint minimizes the validation objective evaluated using that epoch's KL-warmup weight, while the \textit checkpoint minimizes the validation generation score, which combines failure, validity, duplication, training-match, completion, unresolved-stub, and log-scaled structural-MMD terms.
The last completed epoch is reported in the detailed sensitivity tables for diagnostic purposes only and is not considered a third checkpoint candidate.
Test-set MMD and FCD are not used for checkpoint selection.

\subsection{Baseline Models}
The methods included in our comparison span several families of graph generative models.
GraphVAE decodes a continuous latent representation into probabilistic adjacency and attribute matrices~\cite{simonovsky2018graphvae}, while GNF combines a reversible graph neural network with a graph autoencoder to generate each graph in a single pass~\cite{liu2019graph}.
EDP-GNN instead learns a permutation-equivariant score function over adjacency matrices~\cite{niu2020permutation}, while GDSS jointly evolves node features and adjacency variables through a system of stochastic differential equations~\cite{jo2022score}.
CatFlow derives a categorical flow through variational flow matching and predicts a distribution over clean endpoints for each graph component~\cite{eijkelboom2024variational}.

For molecular graphs, MoFlow applies an invertible flow to atom and bond representations~\cite{zang2020moflow}, while GraphEBM learns a permutation-invariant energy function and samples molecules using Langevin dynamics~\cite{liu2021graphebm}.
The molecular comparison further includes diffusion and flow-matching approaches.
DiGress is a discrete denoising diffusion model in which a graph Transformer reconstructs node and edge categories from corrupted graphs~\cite{vignac2023digress}.
Standard Flow Matching learns the vector field defining a continuous normalizing flow~\cite{lipman2023flow}, while Dirichlet Flow Matching defines probability paths directly on the categorical simplex~\cite{stark2024dirichlet}.
For Dirichlet Flow Matching, we report the graph-generation adaptation evaluated alongside CatFlow, rather than the original results on DNA sequences.
DeFoG employs discrete flow matching to generate graphs while decoupling the training and sampling processes, enabling flexible and efficient graph generation~\cite{qin2025defog}.

The PROTEINS comparison additionally includes two autoregressive methods: GraphRNN generates adjacency sequences using recurrent neural networks~\cite{you2018graphrnn}, while GRAN generates graph elements in blocks using recurrent attention~\cite{liao2019efficient}.
It also includes SPECTRE, which first generates the leading eigenvalues and eigenvectors of the graph Laplacian and then conditions a graph generator on these spectral features~\cite{martinkus2022spectre}.
Its ``real spectra'' variant instead uses spectra extracted from test graphs, serving as an oracle diagnostic rather than a fully generative baseline.
The starred MolGAN and GG-GAN variants are architectural controls introduced by the SPECTRE authors, rather than faithful re-implementations of the original methods.
MolGAN* replaces SPECTRE's conditional graph generator with an MLP.
GG-GAN* combines fixed learned node embeddings with processed noise, while its random-set variant uses only the processed random set.

\subsection{Evaluation Details}
For general graph generation, we use the official ordered $80/20$ train--test split from~\cite{eijkelboom2024variational}.
Orbit MMD is computed using the $15$ exact ORCA orbit features corresponding to graphlets of up to four nodes.
For the supplementary SPECTRE experiments, we preserve the fixed dataset membership and generate one graph per test graph.
Each reported ratio is the generated--test $\mathrm{MMD}^2$ divided by the corresponding published train--test $\mathrm{MMD}^2$.

For molecular generation, we use CatFlow's preprocessing pipeline and the official test set.
Validity is computed over all independent generation attempts, uniqueness over valid generated molecules, and FCD over valid generated molecules with duplicates retained.
FCD measures the distance between the distributions of real and generated molecules using activations from ChemNet.

\subsection{Three-Seed Results}\label{app:multiseed-results}
We report the arithmetic mean and population standard deviation over three different seeds.
Tables~\ref{tab:sup-ego-checkpoints}--\ref{tab:sup-molecule-checkpoints} show sensitivity across WL depths and checkpoint
roles, including the last completed epoch as a diagnostic.

\begin{table}[h]
\centering
\scriptsize
\setlength{\tabcolsep}{5pt}
\begin{tabular}{clcccc}
\toprule
WL iterations & Checkpoint & Degree MMD$^2$ & Clustering MMD$^2$ & Orbit MMD$^2$ & Complete (\%) \\
\midrule
1 & Best structural & $0.0246 \pm 0.0145$ & $0.0407 \pm 0.0281$ & $0.0071 \pm 0.0038$ & $84.17 \pm 11.79$ \\
1 & Best val. loss & $0.0239 \pm 0.0124$ & $0.0334 \pm 0.0153$ & $0.0080 \pm 0.0036$ & $89.17 \pm 4.71$ \\
1 & Last epoch (diagnostic) & $0.0255 \pm 0.0141$ & $0.0328 \pm 0.0182$ & $0.0076 \pm 0.0037$ & $90.83 \pm 2.36$ \\
2 & Best structural & $0.0251 \pm 0.0131$ & $0.0373 \pm 0.0241$ & $0.0078 \pm 0.0038$ & $83.33 \pm 5.14$ \\
2 & Best val. loss & $0.0242 \pm 0.0132$ & $0.0329 \pm 0.0199$ & $0.0079 \pm 0.0038$ & $83.33 \pm 3.12$ \\
2 & Last epoch (diagnostic) & $0.0256 \pm 0.0136$ & $0.0336 \pm 0.0208$ & $0.0078 \pm 0.0038$ & $84.17 \pm 2.36$ \\
3 & Best structural & $0.0245 \pm 0.0104$ & $0.0319 \pm 0.0071$ & $0.0095 \pm 0.0019$ & $75.83 \pm 12.47$ \\
3 & Best val. loss & $0.0223 \pm 0.0127$ & $0.0338 \pm 0.0197$ & $0.0078 \pm 0.0037$ & $85.83 \pm 6.56$ \\
3 & Last epoch (diagnostic) & $0.0236 \pm 0.0131$ & $0.0349 \pm 0.0228$ & $0.0078 \pm 0.0037$ & $88.33 \pm 6.24$ \\
\bottomrule
\end{tabular}
\caption{Checkpoint sensitivity results for Ego-Small under the CatFlow evaluation protocol.}
\label{tab:sup-ego-checkpoints}
\end{table}

\begin{table}[h]
\centering
\scriptsize
\setlength{\tabcolsep}{5pt}
\begin{tabular}{clcccc}
\toprule
WL iterations & Checkpoint & Degree MMD$^2$ & Clustering MMD$^2$ & Orbit MMD$^2$ & Complete (\%) \\
\midrule
1 & Best structural & $0.0258 \pm 0.0027$ & $0.3920 \pm 0.0396$ & $0.1071 \pm 0.0279$ & $16.67 \pm 4.71$ \\
1 & Best val. loss & $0.0296 \pm 0.0010$ & $0.2890 \pm 0.0457$ & $0.0620 \pm 0.0361$ & $21.67 \pm 13.12$ \\
1 & Last epoch (diagnostic) & $0.0290 \pm 0.0022$ & $0.4218 \pm 0.0517$ & $0.1305 \pm 0.0682$ & $16.67 \pm 2.36$ \\
2 & Best structural & $0.0215 \pm 0.0044$ & $0.2244 \pm 0.0827$ & $0.0433 \pm 0.0057$ & $20.00 \pm 7.07$ \\
2 & Best val. loss & $0.0324 \pm 0.0031$ & $0.3243 \pm 0.0386$ & $0.0762 \pm 0.0205$ & $16.67 \pm 13.12$ \\
2 & Last epoch (diagnostic) & $0.0249 \pm 0.0065$ & $0.3145 \pm 0.0235$ & $0.0999 \pm 0.0056$ & $20.00 \pm 4.08$ \\
3 & Best structural & $0.0190 \pm 0.0110$ & $0.1839 \pm 0.0609$ & $0.0232 \pm 0.0103$ & $20.00 \pm 7.07$ \\
3 & Best val. loss & $0.0308 \pm 0.0019$ & $0.3347 \pm 0.0092$ & $0.0903 \pm 0.0129$ & $16.67 \pm 10.27$ \\
3 & Last epoch (diagnostic) & $0.0308 \pm 0.0019$ & $0.3347 \pm 0.0092$ & $0.0903 \pm 0.0129$ & $16.67 \pm 10.27$ \\
\bottomrule
\end{tabular}
\caption{Checkpoint sensitivity results for Community-Small under the CatFlow evaluation protocol.}
\label{tab:sup-community-checkpoints}
\end{table}

\begin{table}[h]
\centering
\scriptsize
\setlength{\tabcolsep}{5pt}
\begin{tabular}{clcccc}
\toprule
WL iterations & Checkpoint & Degree MMD$^2$ & Clustering MMD$^2$ & Orbit MMD$^2$ & Complete (\%) \\
\midrule
1 & Best structural & $0.0346 \pm 0.0056$ & $0.5306 \pm 0.0518$ & $0.0263 \pm 0.0149$ & $37.04 \pm 2.64$ \\
1 & Best val. loss & $0.0323 \pm 0.0043$ & $0.5634 \pm 0.0057$ & $0.0401 \pm 0.0048$ & $33.90 \pm 6.29$ \\
1 & Last epoch (diagnostic) & $0.0402 \pm 0.0029$ & $0.5608 \pm 0.0451$ & $0.0289 \pm 0.0133$ & $35.61 \pm 2.64$ \\
2 & Best structural & $0.0408 \pm 0.0196$ & $0.5388 \pm 0.0233$ & $0.0252 \pm 0.0131$ & $34.47 \pm 3.84$ \\
2 & Best val. loss & $0.0365 \pm 0.0075$ & $0.5351 \pm 0.0370$ & $0.0276 \pm 0.0124$ & $37.89 \pm 0.40$ \\
2 & Last epoch (diagnostic) & $0.0365 \pm 0.0075$ & $0.5351 \pm 0.0370$ & $0.0276 \pm 0.0124$ & $37.89 \pm 0.40$ \\
3 & Best structural & $0.0370 \pm 0.0012$ & $0.5268 \pm 0.0259$ & $0.0200 \pm 0.0052$ & $35.04 \pm 1.85$ \\
3 & Best val. loss & $0.0297 \pm 0.0048$ & $0.5711 \pm 0.0474$ & $0.0392 \pm 0.0120$ & $31.91 \pm 0.40$ \\
3 & Last epoch (diagnostic) & $0.0266 \pm 0.0010$ & $0.5269 \pm 0.0218$ & $0.0274 \pm 0.0037$ & $39.03 \pm 5.42$ \\
\bottomrule
\end{tabular}
\caption{Checkpoint sensitivity results for Enzymes under the CatFlow evaluation protocol.}
\label{tab:sup-enzymes-checkpoints}
\end{table}

\begin{table}[h]
\centering
\scriptsize
\setlength{\tabcolsep}{5pt}
\begin{tabular}{clcccc}
\toprule
WL iterations & Checkpoint & Degree MMD$^2$ & Clustering MMD$^2$ & Orbit MMD$^2$ & Complete (\%) \\
\midrule
1 & Best structural & $0.1583 \pm 0.0711$ & $1.0976 \pm 0.5431$ & $0.4842 \pm 0.2780$ & $31.67 \pm 4.71$ \\
1 & Best val. loss & $0.1738 \pm 0.0536$ & $0.2014 \pm 0.0158$ & $0.1117 \pm 0.0269$ & $16.67 \pm 6.24$ \\
1 & Last epoch (diagnostic) & $0.1738 \pm 0.0536$ & $0.2014 \pm 0.0158$ & $0.1117 \pm 0.0269$ & $16.67 \pm 6.24$ \\
2 & Best structural & $0.1485 \pm 0.0276$ & $0.2730 \pm 0.0716$ & $0.1362 \pm 0.0465$ & $20.00 \pm 0.00$ \\
2 & Best val. loss & $0.1984 \pm 0.0459$ & $0.1895 \pm 0.0714$ & $0.0951 \pm 0.0146$ & $11.67 \pm 13.12$ \\
2 & Last epoch (diagnostic) & $0.1984 \pm 0.0459$ & $0.1895 \pm 0.0714$ & $0.0951 \pm 0.0146$ & $11.67 \pm 13.12$ \\
3 & Best structural & $0.0796 \pm 0.0202$ & $0.1980 \pm 0.1234$ & $0.0649 \pm 0.0368$ & $6.67 \pm 6.24$ \\
3 & Best val. loss & $0.1324 \pm 0.0416$ & $0.1174 \pm 0.0782$ & $0.0442 \pm 0.0164$ & $11.67 \pm 8.50$ \\
3 & Last epoch (diagnostic) & $0.1324 \pm 0.0416$ & $0.1174 \pm 0.0782$ & $0.0442 \pm 0.0164$ & $11.67 \pm 8.50$ \\
\bottomrule
\end{tabular}
\caption{Checkpoint sensitivity results for Grid under the CatFlow evaluation protocol.}
\label{tab:sup-grid-checkpoints}
\end{table}

\begin{table}[h]
\centering
\scriptsize
\setlength{\tabcolsep}{5pt}
\begin{tabular}{llcccc}
\toprule
Dataset & WL iterations & Checkpoint & Validity (\%) & Uniqueness (\%) & FCD \\
\midrule
QM9 & 1 & Best structural & $99.72 \pm 0.06$ & $94.86 \pm 0.26$ & $1.5698 \pm 0.0470$ \\
QM9 & 1 & Best val. loss & $99.76 \pm 0.09$ & $95.02 \pm 0.67$ & $1.8623 \pm 0.4472$ \\
QM9 & 1 & Last epoch (diagnostic) & $99.72 \pm 0.00$ & $95.18 \pm 0.15$ & $1.5511 \pm 0.0435$ \\
QM9 & 2 & Best structural & $99.73 \pm 0.08$ & $94.77 \pm 0.26$ & $1.5702 \pm 0.0498$ \\
QM9 & 2 & Best val. loss & $99.74 \pm 0.05$ & $94.74 \pm 0.42$ & $1.4071 \pm 0.0793$ \\
QM9 & 2 & Last epoch (diagnostic) & $99.70 \pm 0.04$ & $94.61 \pm 0.17$ & $1.5190 \pm 0.0215$ \\
QM9 & 3 & Best structural & $99.79 \pm 0.07$ & $94.91 \pm 0.39$ & $1.8669 \pm 0.4918$ \\
QM9 & 3 & Best val. loss & $99.71 \pm 0.06$ & $94.61 \pm 0.09$ & $1.4103 \pm 0.0265$ \\
QM9 & 3 & Last epoch (diagnostic) & $99.66 \pm 0.11$ & $94.78 \pm 0.11$ & $1.4555 \pm 0.0368$ \\
\midrule
ZINC250K & 2 & Best structural & $93.16 \pm 1.58$ & $100.00 \pm 0.00$ & $18.1730 \pm 0.1242$ \\
ZINC250K & 2 & Best val. loss & $90.49 \pm 1.10$ & $100.00 \pm 0.00$ & $17.9320 \pm 0.0951$ \\
ZINC250K & 2 & Last epoch (diagnostic) & $90.68 \pm 1.15$ & $100.00 \pm 0.00$ & $17.8237 \pm 0.0136$ \\
ZINC250K & 3 & Best structural & $93.34 \pm 1.17$ & $100.00 \pm 0.00$ & $18.6496 \pm 0.4463$ \\
ZINC250K & 3 & Best val. loss & $88.22 \pm 2.22$ & $100.00 \pm 0.00$ & $17.9404 \pm 0.0682$ \\
ZINC250K & 3 & Last epoch (diagnostic) & $87.65 \pm 1.95$ & $100.00 \pm 0.00$ & $17.8087 \pm 0.2409$ \\
\bottomrule
\end{tabular}
\caption{Checkpoint sensitivity results for QM9 and ZINC250K under the CatFlow evaluation protocol. QM9 uses the heavy-atom representation.}
\label{tab:sup-molecule-checkpoints}
\end{table}

\subsection{Additional Results}\label{app:baseline-comparisons}

\paragraph{Proteins Dataset.}
We additionally evaluate WL-SeqVAE on PROTEINS using the SPECTRE evaluation protocol.
Among the validation-selected candidates, WL with $2$ iterations and the best-structural checkpoint is the only configuration that generates all $184$ required graphs for every seed.
Table~\ref{tab:sup-proteins-spectre} compares this configuration with the published SPECTRE baselines.
As this protocol defines no dataset-specific validity predicate for PROTEINS, we report uniqueness and novelty directly.

\begin{table}[t]
\centering
\scriptsize
\setlength{\tabcolsep}{2pt}
\renewcommand{\arraystretch}{1.08}
\begin{tabular}{@{}lc|c|c|ccc@{}}
\toprule
& \multicolumn{6}{c}{PROTEINS} \\
\cmidrule(lr){2-7}
Model & Deg. \(\downarrow\) & Clus. \(\downarrow\) & Orbit \(\downarrow\)
& Unique \(\uparrow\) & Novel \(\uparrow\) & Uniq. \& Nov. \(\uparrow\)\\
\midrule
Training set
& 0.0003 & 0.0068 & 0.0032 
& 100.0 & -- & -- \\
GraphRNN
& 0.0040 & 0.1475 & 0.5851 
& 100.0 & 100.0 & 100.0 \\
GRAN
& 0.0479 & 0.1234 & 0.3458
& 100.0 & 100.0 & 100.0 \\
MolGAN\(^{*}\)
& 0.0008 & 0.0644 & 0.0081
& 97.3 & 100.0 & 97.3 \\
GG-GAN (RS)\(^{*}\)
& 0.4727 & 0.1772 & 0.7326
& 100.0 & 100.0 & 100.0 \\
GG-GAN\(^{*}\)
& 0.5192 & 0.5220 & 0.7326
& 100.0 & 100.0 & 100.0 \\
SPECTRE (\(k=16\))
& 0.0056 & 0.0843 & \textbf{0.0267}
& 100.0 & 100.0 & 100.0 \\
\shortstack[l]{SPECTRE (\(k=16\),\\ real spectra)}
& \textbf{0.0013} & \textbf{0.0469} & 0.0287
& 100.0 & 100.0 & 100.0\\
\midrule
\textbf{WL-SeqVAE}
& 0.0221 {\tiny$\pm$  0.01}
& 0.1523 {\tiny $\pm$ 0.03}
& 1.0738 {\tiny $\pm$ 0.06}
& 100.0 
& 100.0 
& 100.0 \\
\bottomrule
\end{tabular}
\caption{Graph generation results on PROTEINS under the SPECTRE evaluation protocol. Baseline results are taken from~\cite{martinkus2022spectre}. WL-SeqVAE values are reported as the mean $\pm$ standard deviation over different runs. The best results are highlighted in bold.}
\label{tab:sup-proteins-spectre}
\end{table}

\begin{table}[!h]
\centering
\footnotesize
\setlength{\tabcolsep}{2pt}
\begin{tabular}{@{}lcccc@{}}
\toprule
Model
& Deg. $\downarrow$
& Clus. $\downarrow$
& Orb. $\downarrow$
& V.U.N. $\uparrow$ \\
\midrule

\multicolumn{5}{c}{\textit{Stochastic Block Model}} \\
\midrule

SPECTRE
    & 1.9
    & 1.6
    & 1.6
    & 53\% \\

ConGress
    & 34.1
    & 3.1
    & 4.5
    & 0\% \\

DiGress
    & 1.6
    & \textbf{1.5}
    & 1.7
    & 74\% \\

Flow Matching
    & 10.2
    & 2.0
    & 3.2
    & 22\% \\

Dirichlet FM
    & 1.7
    & \textbf{1.5}
    & \textbf{1.4}
    & \textbf{80\%} \\

CatFlow
    & \textbf{1.5}
    & \textbf{1.5}
    & \textbf{1.4}
    & \textbf{85\%} \\

\midrule

\textbf{WL-SeqVAE}
    & 2.5 $\pm$ {\scriptsize 0.45} 
    & 1.86 $\pm$ {\scriptsize 0.04} 
    & 4.10 $\pm$ {\scriptsize 0.59} 
    & 7.4\% \\

\midrule

\multicolumn{5}{c}{\textit{Planar Graphs}} \\
\midrule

SPECTRE
    & 2.5
    & 2.5
    & 2.4
    & 25\% \\

ConGress
    & 23.8
    & 8.8
    & 2590
    & 0\% \\

DiGress
    & \textbf{1.4}
    & \textbf{1.2}
    & 1.7
    & 75\% \\

Flow Matching
    & 5.1
    & 5.6
    & 5.5
    & 30\% \\

Dirichlet FM
    & 1.5
    & 1.3
    & \textbf{1.5}
    & 80\% \\

CatFlow
    & \textbf{1.4}
    & 1.3
    & \textbf{1.5}
    & 80\% \\

\midrule

\textbf{WL-SeqVAE}
    & 109.5 $\pm$ {\scriptsize 10.6}
    & 11.2  $\pm$ {\scriptsize 0.1}
    & 92.4 $\pm$ {\scriptsize 17.5 }
    & \textbf{100\%} \\

\bottomrule
\end{tabular}
\caption{
Additional Graph-Generation Results on the Stochastic Block Model (SBM) and Planar Graph Datasets. Lower values are better for degree (\textit{Deg.}), clustering (\textit{Clus.}), and orbit (\textit{Orb.}) statistics, while higher values are better for Validity, Uniqueness, and Novelty (V.U.N.).
The best results are highlighted in bold.
}
\label{tab:sup-planar-sbm-baselines}
\end{table}

WL-SeqVAE achieves a lower degree $\mathrm{MMD}^2$ than GRAN and both GG-GAN variants, but is outperformed by GraphRNN, MolGAN*, and both SPECTRE configurations
Its clustering $\mathrm{MMD}^2$ is comparable to that of GraphRNN and lower than those of both GG-GAN variants.
However, its orbit $\mathrm{MMD}^2$ is higher than that of every reported baseline.
Despite these structural discrepancies, WL-SeqVAE generates a fully unique and novel collection of graphs.
Overall, the PROTEINS results suggest that WL-SeqVAE preserves sample diversity but does not yet match SPECTRE in reproducing the higher-order structural statistics of the reference distribution.

\paragraph{Synthetic Graph Generation.}
We evaluate Planar and SBM using the official $200$-graph SPECTRE datasets and the DiGress split of $128$ training, $32$ validation, and $40$ test graphs.
For both datasets, we use $3$ WL iterations and the best-structural checkpoint for each of the three training seeds.
Each run is expected to generate $40$ graphs, and we report aggregate results across seeds only when every run produces all $40$ graphs. Table~\ref{tab:sup-planar-sbm-baselines} compares WL-SeqVAE with the baselines.
The reported ratios are computed by dividing each generated--test $\mathrm{MMD}^2$ by the corresponding published train--test $\mathrm{MMD}^2$.

All three Planar runs produce the required $40$ graphs.
WL-SeqVAE achieves $100\%$ V.U.N., compared with $80\%$ for CatFlow, indicating that every generated graph satisfies the evaluator's requirements for planarity, connectivity, uniqueness, and novelty.
Nevertheless, its degree, clustering, and orbit ratios are substantially higher than those of the baselines, demonstrating that satisfying dataset-specific validity constraints does not necessarily imply close agreement with the reference distribution.
The $0\%$ strict WL-completion rate is not contradictory since a graph may satisfy the dataset validity criteria while leaving one or more sampled local WL requirements unresolved.

None of the SBM runs produces all $40$ requested graphs.
We therefore report results separately for each seed and refrain from ranking their ratios against those of the complete baseline collections.
Nevertheless, the low V.U.N. and comparatively high orbit ratios suggest that local WL compatibility alone is insufficient to reproduce the global community structure characteristic of SBM graphs.
These results motivate incorporating community-level or spectral conditioning.

\subsection{Vocabulary Growth and Dataset-Dependent Training Time}
\begin{figure*}[h]
\centering
\includegraphics[width=\textwidth]{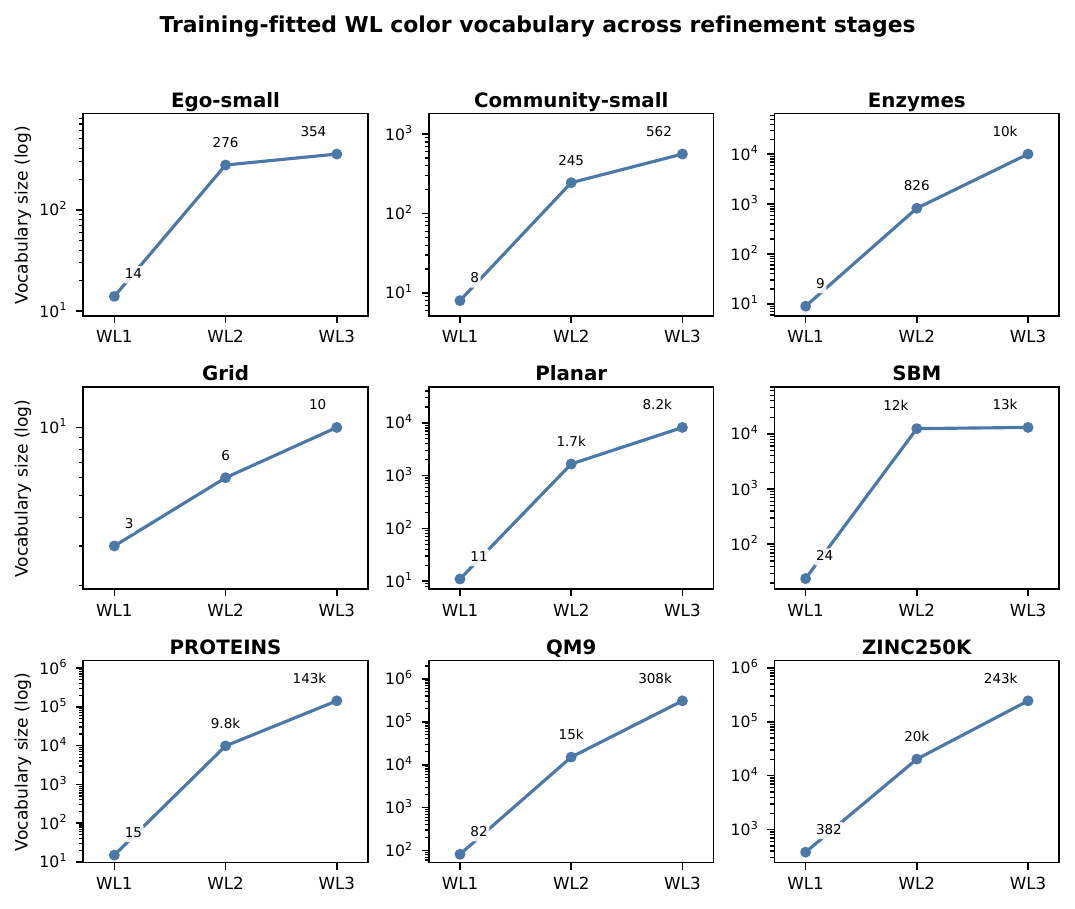}
\caption{Growth of the WL-Color vocabulary fitted on the training set from $1$ iteration to $3$ iterations. Vocabulary sizes are identical across seeds. For datasets without a checkpoint at a given depth, the vocabulary size comes from the intermediate dictionaries stored in a deeper WL trace. Each panel uses a dataset-specific logarithmic vertical scale.}
\label{fig:sup-wl-vocabulary-growth}
\end{figure*}

\begin{figure*}[h]
\centering
\includegraphics[width=\textwidth]{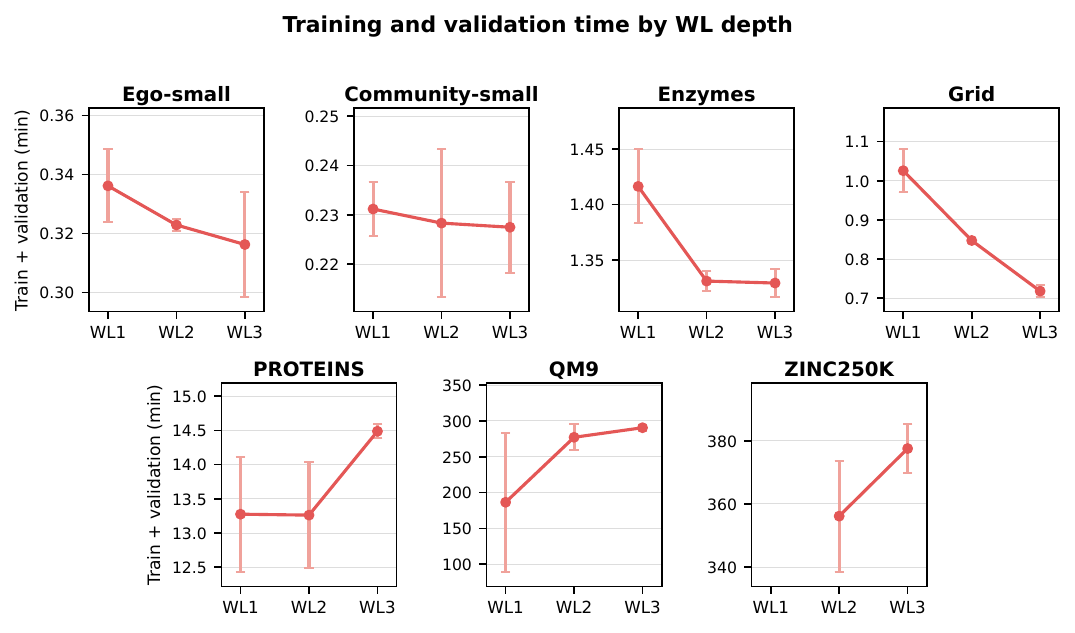}
\caption{Training and validation time from $1$ WL iteration to $3$ iterations. Points and error bars show the mean and standard deviation over three seeds. Each panel uses a dataset-specific linear scale. Only results for $2$ and $3$ iterations are available for ZINC250K. The times exclude trace preprocessing, structural-checkpoint evaluation, and final test generation.}
\label{fig:sup-training-time-by-wl}
\end{figure*}

Figures~\ref{fig:sup-wl-vocabulary-growth} and \ref{fig:sup-training-time-by-wl} illustrate the effects of WL refinement depth on vocabulary size and computational cost.
As shown in Figure~\ref{fig:sup-wl-vocabulary-growth}, the number of distinct WL colors increases substantially with refinement depth, particularly for larger datasets such as QM9, ZINC250K, and PROTEINS.
However, this growth does not translate into a proportional increase in training time.
Figure~\ref{fig:sup-training-time-by-wl} shows that training and validation times remain relatively stable or even decrease on smaller datasets, while larger datasets exhibit more moderate increases.
These results suggest that deeper WL refinement can capture richer structural information without necessarily incurring substantial additional computational overhead.

\section{Samples from WL-SeqVAE}\label{app:generated-examples}
\begin{figure*}[ht]
\centering
\includegraphics[width=\textwidth]{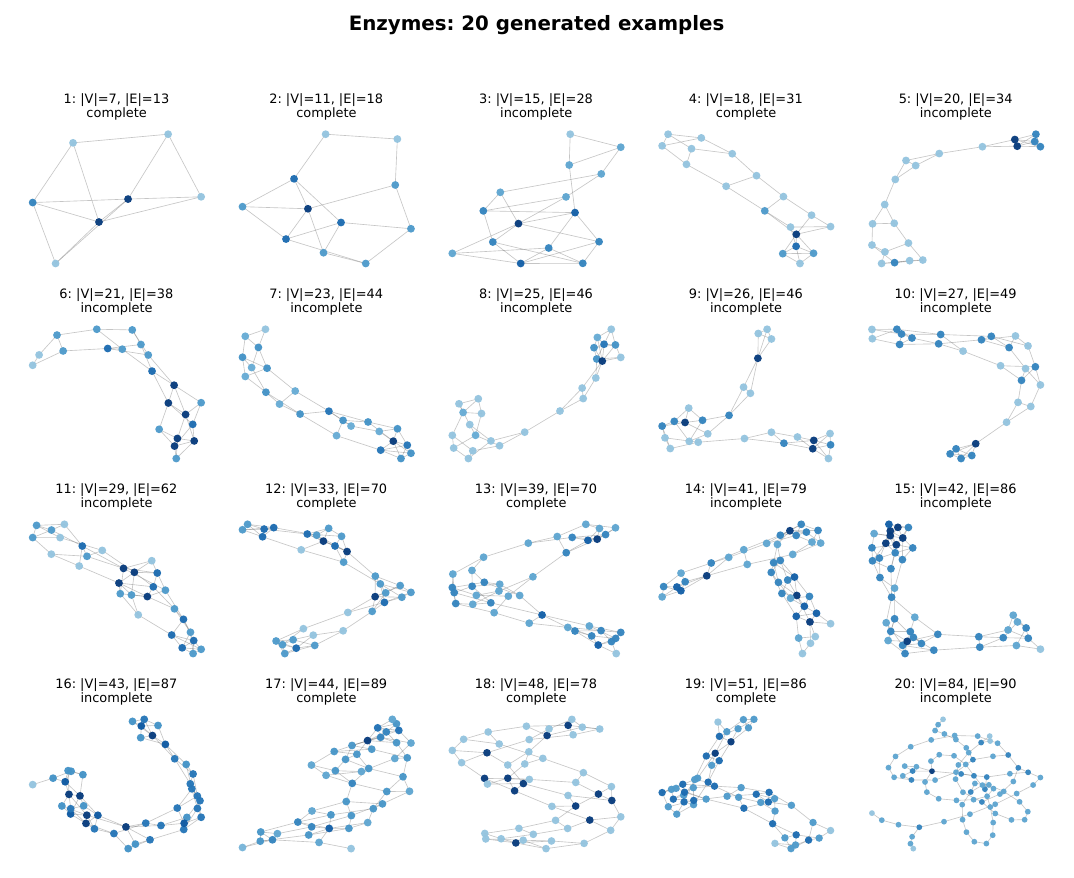}
\caption{$20$ ENZYMES graphs generated by the proposed model. Node color indicates degree.}
\label{fig:sup-generated-enzymes}
\end{figure*}

\begin{figure*}[ht]
\centering
\includegraphics[width=\textwidth]{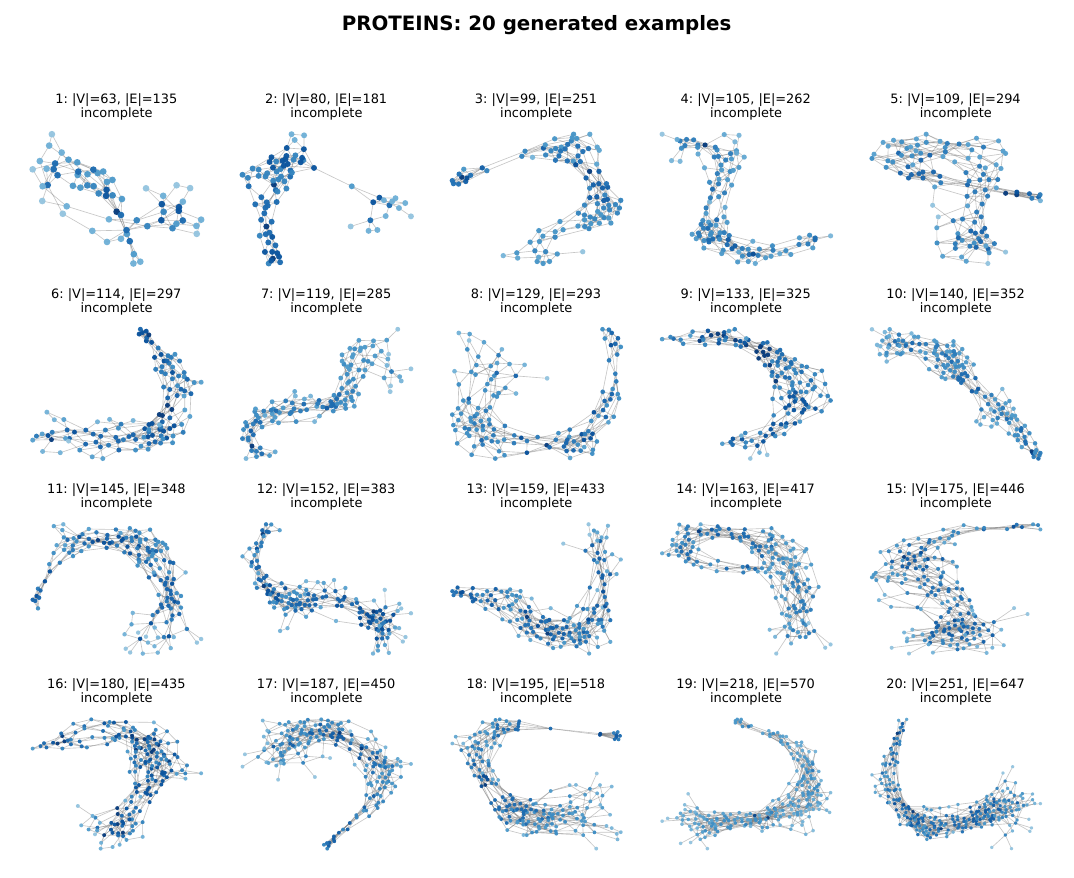}
\caption{$20$ PROTEINS graphs generated by the proposed model. Node color indicates degree.}
\label{fig:sup-generated-proteins}
\end{figure*}

\begin{figure*}[ht]
\centering
\includegraphics[width=\textwidth]{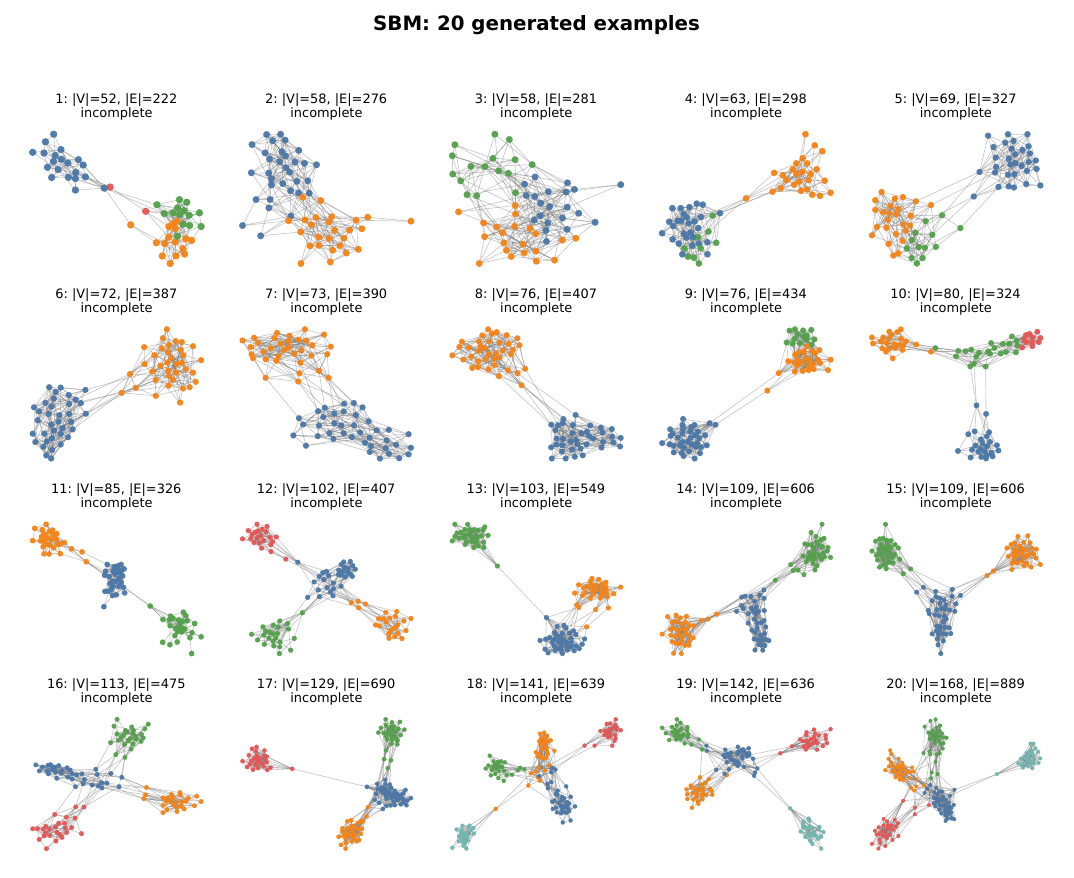}
\caption{$20$ SBM graphs generated by the proposed model. Colors indicate communities inferred by greedy modularity for visualization.}
\label{fig:sup-generated-sbm}
\end{figure*}

\begin{figure*}[ht]
\centering
\includegraphics[width=\textwidth]{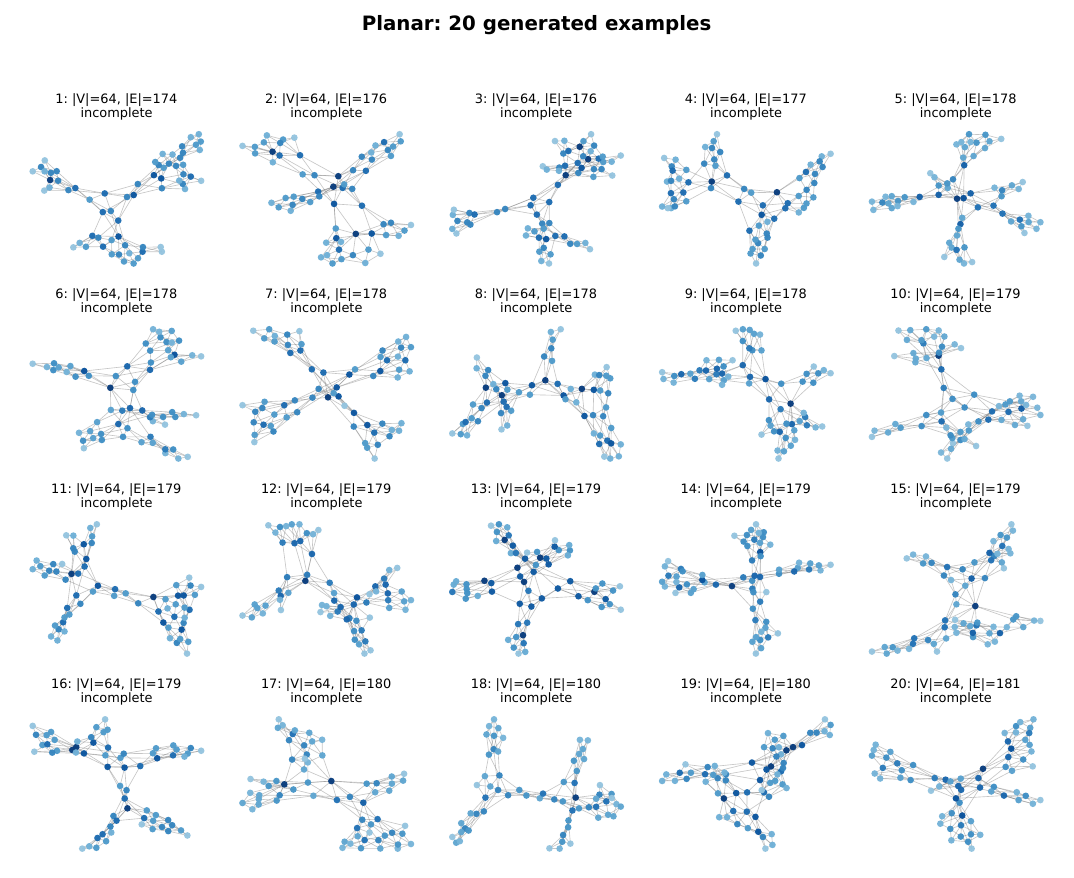}
\caption{$20$ Planar graphs generated by the proposed model. Node color indicates degree.}
\label{fig:sup-generated-planar}
\end{figure*}

Figures~\ref{fig:sup-generated-enzymes}--\ref{fig:sup-generated-planar} show $20$ generated examples for Enzymes, PROTEINS, SBM, and Planar.
For each dataset, we sort the outputs by node count, edge count, edge-list signature, and generation order, then select $20$ evenly spaced examples to represent the range of generated graphs without manual curation.

\begin{figure*}[ht]
\centering
\includegraphics[width=0.87\textwidth]{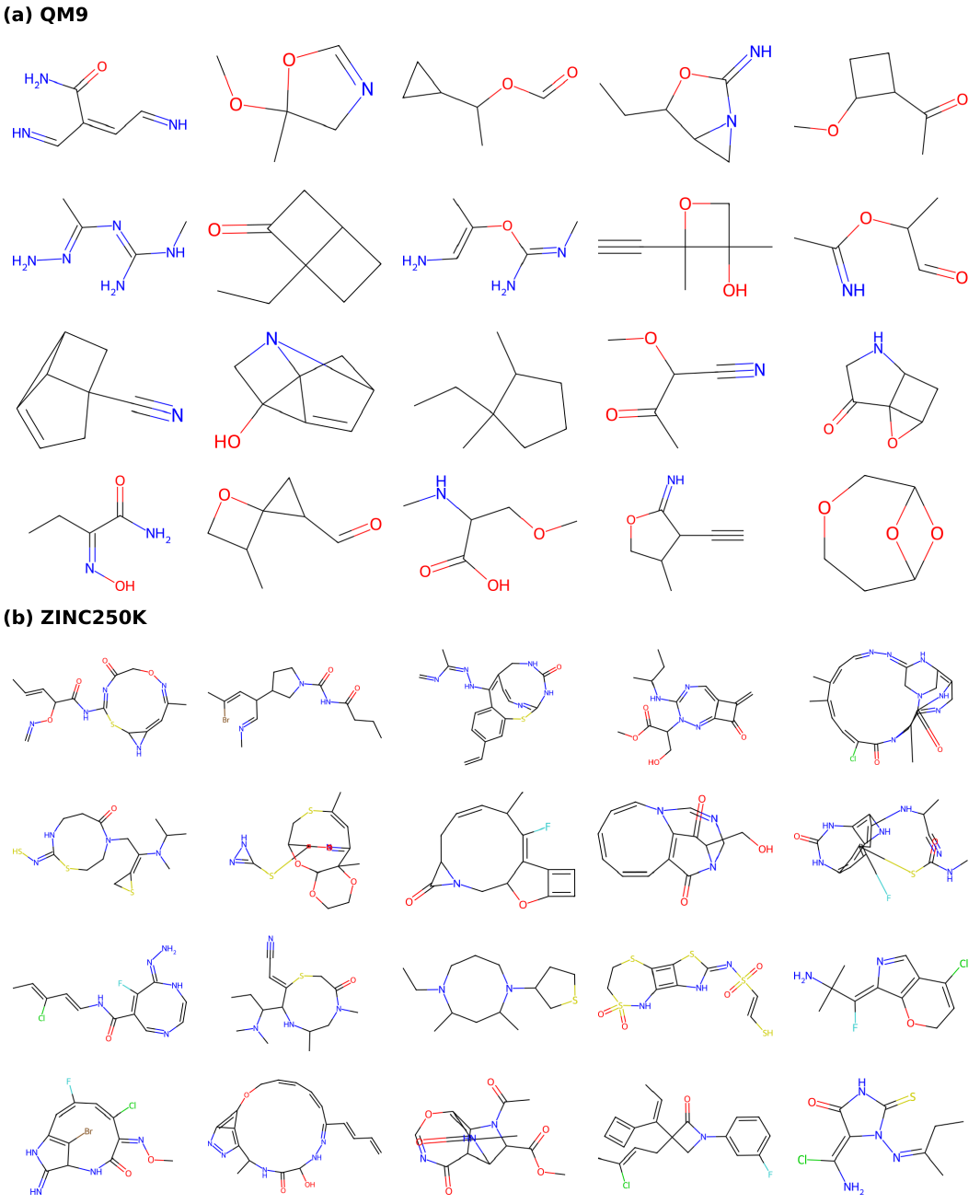}
\caption{$20$ (a) QM9 Molecules and (b) ZINC250K Molecules generated by the proposed model. Each panel contains $20$ Valid, Unique molecules selected at evenly spaced positions in first-occurrence generation order, without manual curation. Standard RDKit atom colors are used. Hydrogens are implicit in the heavy-atom representation.}
\label{fig:sup-generated-molecules}
\end{figure*}

Figure~\ref{fig:sup-generated-molecules} presents the generated molecular examples.
For each dataset, we use the outputs from the validation-selected configuration with $2$ WL iterations and retain valid, unique canonical SMILES in their order of first occurrence.
We then select $20$ evenly spaced examples, again without manual curation.

\section{Limitations}
WL-SeqVAE does not model arbitrary node or edge feature vectors.
The molecular experiments use finite atom and bond categories as discrete labels in WL refinement and decoding.
We do not consider continuous attributes, mixed feature vectors, or conditional graph-level targets.

\subsection{Methodological Limitations}
The main limitation is that generation does not always reach strict WL
completion.
The decoder may stop while one or more sampled WL requirements remain unresolved.
Across the three-seed CatFlow general graph generation results, completion ranges from $11.67\%$ for Grid at $3$ WL iterations to $89.17\%$ for Ego-small at $1$ WL iteration.
Each accepted action is WL-compatible, but the sampler does not guarantee that all local requirements will be completed.

The WL constraints are local.
They ensure reciprocal compatibility between edge endpoints, but they do not guarantee planarity, exact grid structure, community structure, synthesizability, or other molecular properties.
A model can therefore generate connected graphs and chemically valid molecules while performing poorly on datasets with strong global constraints.

\subsection{Practical Limitations}
The decoder is sequential, and the trace vocabulary grows with WL depth
and graph diversity.
Higher WL depths can improve structural specificity, but they also enlarge the color vocabulary and increase sparsity.
Larger molecules or graphs may require longer maximum sequences and larger trace caches to achieve high completion.

\end{document}